\documentclass{article}
\usepackage{arxiv}
\usepackage[utf8]{inputenc} % allow utf-8 input
\usepackage[T1]{fontenc}    % use 8-bit T1 fonts
\usepackage{hyperref}       % hyperlinks
\usepackage{url}            % simple URL typesetting
\usepackage{booktabs}       % professional-quality tables
\usepackage{amsfonts}       % blackboard math symbols
\usepackage{nicefrac}       % compact symbols for 1/2, etc.
\usepackage{microtype}      % microtypography
\usepackage{import}
\usepackage{todonotes}
\usepackage{enumitem}
\usepackage{csquotes}
\usepackage{textcomp}       % for \textpm +-
\usepackage{amsmath}        % for math, duh
\usepackage{gb4e}\noautomath% for linguistic example
\title{Exploring the Combination of Contextual Word Embeddings and Knowledge Graph Embeddings}

\author{
    Lea Dieudonat\thanks{By alphabetical order of authors' last names, all authors are first authors.} \\
    IDMC, Université de Lorraine \\
    Nancy, France \\
    \And
    Kelvin Han\footnotemark[1] \\
    IDMC, Université de Lorraine \\
    Nancy, France \\\\
    \And
    Phyllicia Leavitt\footnotemark[1]\\
    IDMC, Université de Lorraine \\
    Nancy, France \\
    \And
    Esteban Marquer\footnotemark[1] \\
    IDMC, Université de Lorraine \\
    Nancy, France \\
}

\begin{document}
\maketitle

\begin{abstract}
\textquote{Classical} word embeddings, such as Word2Vec, have been shown to capture the semantics of words based on their distributional properties. However, their ability to represent the different meanings that a word may have is limited. Such approaches also do not explicitly encode relations between entities, as denoted by words. Embeddings of knowledge bases (KB) capture the explicit relations between entities denoted by words, but are not able to directly capture the syntagmatic properties of these words. To our knowledge, recent research have focused on representation learning that augment the strengths of one with the other. In this work, we begin exploring another approach using contextual and KB embeddings jointly at the same level and propose two tasks -- an entity typing and a relation typing task -- that evaluate the performance of contextual and KB embeddings. We also evaluated a concatenated model of contextual and KB embeddings with these two tasks, and obtain conclusive results on the first task. We hope our work may contribute as a basis for models and datasets that develop in the direction of this approach. 

\end{abstract}

% keywords can be removed

\section{Introduction\label{sec:intro}}

Many methods in Natural Language Processing (NLP) rely on embeddings to process text and knowledge, particularly methods involving the use of deep learning. The quality of embeddings have been shown to significantly improve the results of a wide range of downstream tasks, and a lot of effort have been made towards the development and training of embedding models during the past two decades.

There are two major trends in the literature on embeddings currently: textual embeddings \cite{wang_etal:2019:word_embeddings,chang2019does,smith2019contextual,tenney2019you,ethayarajh2019contextual,torabi_etal:2018:word_embeddings,bakarov:2018:review_word_embeddings,camacho2018word,peters2018deep,NeelakantanSPM15,mikolov2013efficient} and more recently knowledge embeddings \cite{wang_etal:2017:kg_embeddings,nickel:2016:kg_emb_review,trouillon_etal:2016:complex,lin:2015:transR,wang_etal:2014:transH,yang:2014:distmult,bordes_etal:2013:transE,nickel:2013:rescal}.
Also, in the last three years, attempts have been made to augment textual and contextual embeddings with knowledge \cite{peters_etal:2019:knowledge_enhanced_contextual,sun2019ernie,yu2014improving}, or knowledge embeddings with textual information \cite{ramprasad2019coke,simov:2017:extrinseque_common}.
However, most of these works are recent and focus on augmenting either of the two types of embeddings with additional information from the other. In this manner, there is one kind of information (textual or knowledge) which stays at the core of the model and serves as the input in practice, while the other is used to improve the internal representation of the model. To our knowledge, no results has been published of approaches embedding knowledge and contextual information together, using both as the input to the model (instead of augmenting one with the other), and no evaluation method has been defined that is comparable in the same way for both knowledge and textual embeddings.

We aim to address this gap in the literature by proposing a set of tasks to evaluate the performance of embedding models, designed in a way that knowledge and textual information can be used on the same level.
Such evaluation methods are important towards the development of embedding models using both types of information as input.

First of all we will provide an overview of the existing embedding methods and the techniques for evaluating them (\autoref{sec:relatedwork}, \autopageref{sec:relatedwork}). We will then detail how we approach the problem (\autoref{sec:approach}, \autopageref{sec:approach}) and how we adapt existing datasets to our needs (\autoref{sec:data}, \autopageref{sec:data}). Finally, we will describe our experimental setup (\autoref{sec:experiments}, \autopageref{sec:experiments}), the results (\autoref{sec:results}, \autopageref{sec:results}) we obtain and potential future work (\autoref{sec:futurework}, \autopageref{sec:futurework}).

\section{Related work\label{sec:relatedwork}}

In this section we outline the major trends in the literature on embedding models for text and knowledge bases. We discuss some techniques that have been developed to combine these resources. Finally, we elaborate on the way to evaluate such models.

%\todo{Literature review: Combined Embedding [Phlee + Lea] + KB [Kelvin]
%Tasks entity types + relation prediction [Esteban]
%Evaluation - similarity + relatedness (MAP, MRR, etc…) [Kelvin]}
\subsection{Embedding text}
Machine Learning methods are widely used in NLP problems such as question answering, search engines, and human-machine interaction. Because Machine Learning models require numerical input, it is necessary to transpose textual data into numerical values when treating NLP problems. Given a vocabulary of a language, a possible approach is to assign singular numerical values, or to index the vocabulary. This very simple solution would not, however provide information about the similarity, relatedness, and distributional properties of words \cite{smith2019contextual}. One approach to capturing this information is through the use of word embeddings.

\subsubsection{``Static'' word embeddings}
Word embeddings -- which may also be called vector-space models or  distributional semantics models -- are the representation of words as vectors. Distributional semantics corresponds to a concept in linguistics first introduced by Wittgenstein in his work, Philosophical Investigations, arguing that a word may be represented by the way in which it is used \cite{wittgenstein1953philosophical}. This, along with the idea that language has a distributional structure \cite{harris1954distributional}, has been  the driving factor behind the development of word embeddings, which are derived from a probability distribution of words found in their context, obtained by training on large, unlabeled textual corpus. The approaches to training word embeddings may be broken into three broad categories: pre-neural, neural and contextualized.

In the classical approach, pre-neural models output word embedding by applying Latent Semantic Analysis (LSA) to a term-document matrix, making use of matrix dimension reduction by means of Singular Value Decomposition (SVD), which is a matrix decomposition operation.  

Recently, the development of neural techniques for training word embeddings has made an impact on the NLP community. Namely, the model developed by \cite{mikolov2013efficient} made available embeddings pre-trained on large corpora by means of neural network techniques. While there are different techniques for encoding them, traditional, ``static'' embeddings
such as Word2Vec, Glove and fastText represent words by a single vector per token of a vocabulary. 

Although these embeddings encode what we have previously referred to as \textquote{distributional semantics}, they do not capture per se, information such as hyponymy, synonymy, and other linguistic and/or factual information which is present in linguistic resources \cite{ramprasad2019coke}.

%\newline

\subsubsection{Contextual word embeddings\label{subsec:relatedwork-contword_emb}}

Contextual word representations was first introduced by \cite{peters2018deep}, and allows the encoding of embeddings which vary \textquote{dynamically} with the input sequence. For example, if given a sentence \blockquote{John and his friends play soccer}, and \blockquote{I attended the play}, there would be different output for the polysemous token \blockquote{play}. Such embeddings are achieved by training language models and outputting not only the word embeddings themselves, but also the weights that are achieved through training\cite{smith2019contextual}. Significant improvements have been observed on NLP tasks which use contextualized embeddings as input \cite{peters2018deep}.

Since the introduction of ELMo in 2018, other contextual embedding models have been developed. While they differ in the neural network architectures and the language modeling tasks they are trained on, all these models have in common the ability to represent words in context. What is actually learned in each of these embedding models is however an open topic of research\cite{ethayarajh2019contextual}. Although \cite{peters2018deep} observe that the word vectors generated from the first Bi-LSTM\footnote{ELMo's architecture comprises a three-layer structure where layer zero is the character-based context independent layer, followed by two Bi-directional Long-Short Term Memory (Bi-LSTM) layers.} layer seem to better capture syntax and the second layer seem to capture semantics better.

%%%%%%%%%%%%%%%%%%%%%%%%%%%%%%%%%%%%%%%%
\subsection{Knowledge graph embeddings\label{subsec:relatedwork-kg_emb}}

KBs are a store of information for use by computer systems. KBs with structured, relational information are typically referred to as knowledge graphs (KGs). They contain relational information between entities, as well as the properties of these entities and their relations. Thanks to the relative scalability afforded by crowd-sourcing, very large KGs have been built. These include Freebase (see \autoref{sec:data} \autopageref{sec:data}) and Google's Knowledge Graph, which typically store information about entities and the relations between them as triple with the following form \texttt{<subject, relation, object>}. Typically, KB entities include nominals as well as abstract concepts, whereas KB relations may have one or more of the following properties: symmetry, reflexivity, transitivity and equivalence.  

However, because of the symbolic nature of their representation and their sizes, KGs are difficult to manipulate\cite{wang_etal:2017:kg_embeddings}. Within the last decade, new approaches have been proposed to represent the discrete symbolic information in KGs in continuous vector spaces, by embedding entities and relations in these spaces. Two broad classes of such KG embeddings approaches have emerged - namely, translation-based models (\autoref{subsubsec:relatedwork-kg_emb-trans}) and latent feature models (\autoref{subsubsec:relatedwork-kg_emb-latent}). 

\subsubsection{Translation models\label{subsubsec:relatedwork-kg_emb-trans}}

\begin{figure}
    \centering
    \includegraphics[scale=0.5]{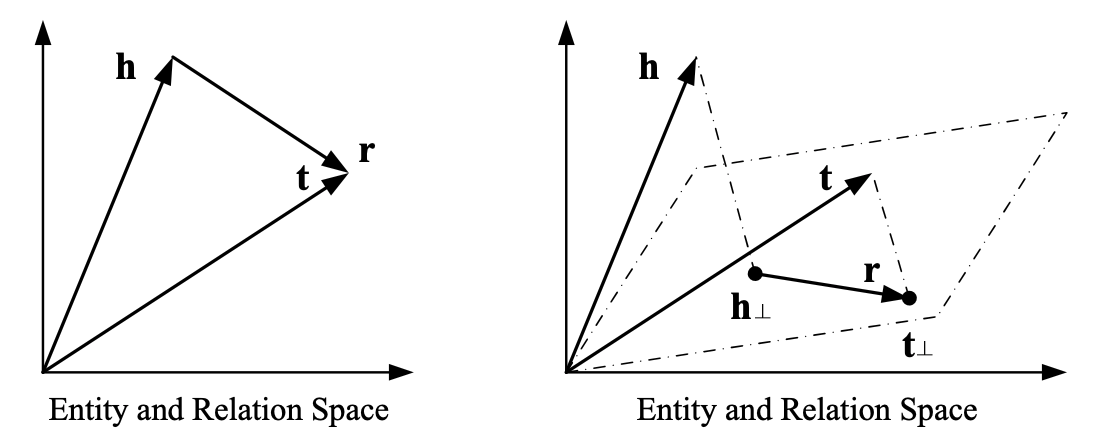}
    \caption{Visualisations of translational KG embedding models in lower-dimensional space. The figure on the left illustrates the TransE approach to embedding entities and relations, the figure on the right illustrates the same for the TransH model. Source: \cite{wang_etal:2017:kg_embeddings}}
    \label{fig:translational}
\end{figure}

The \texttt{TransE} model proposed in \cite{bordes_etal:2013:transE} is the earliest member of the translational class of KG embedding models. It works by modelling the relation between a head entity and a tail entity as a translation in vector space, and embeds KG information by optimising the positions of entities and relations globally across the embedding space. \autoref{fig:translational} (\autopageref{fig:translational}) includes a depiction of how entities and relations are represented in the TransE model.  

The TransE model was however found to be too restrictive in representing relations. While it is suitable for embedding KGs with one-to-one relations, it is unable to sufficiently embed the full properties of KGs that contain one-to-many, many-to-one and many-to-many relations \cite{wang_etal:2017:kg_embeddings,wang_etal:2014:transH}. Models extending the capabilities of TransE have been proposed, these include  \texttt{TransH} \cite{wang_etal:2014:transH} and \texttt{TransR} \cite{lin:2015:transR}, as well as more recent models which seek to extend specific aspects of each of these three models. 

TransH, proposed in \cite{wang_etal:2014:transH} casts the representation of each relation as a hyperplane in the vector space. Relations between entities are then translations across these hyperplanes. In doing so, TransH expands the capability of TransE to represent multiple relations between entities. An illustration of the approach taken in TransH can be found in \autoref{fig:translational} (\autopageref{fig:translational}). 

The TransR model in \cite{lin:2015:transR} adopts a different approach by embedding relations and entities in separate vector spaces. Specifically, each relation in TransR is represented by a projection matrix. An application of the projection matrix of relation \textit{r} on the entities embedding space will project the entities onto \textit{r}'s embedding space. Therefore, a translation of \textit{r} on the head entity in this r-projected space will identify tail entities it is related to by \textit{r}, within a neighbourhood of the translation. In doing so, TransE extends the representational power of TransH by allow translations beyond hyperplanes, in higher dimensional vector spaces. 

While these extensions increase the expressivity of the initial TransE model, the expressivity comes at the expense of higher complexity \cite{wang_etal:2017:kg_embeddings}. For instance, TransR requires a projection matrix for each relation in the KG, thereby increasing significantly the training and memory requirement for the model. 

\subsubsection{Latent feature models \label{subsubsec:relatedwork-kg_emb-latent}}
Another class of KG embedding models approaches the problem by representing KG information in the form of a tensor\footnote{Such KG tensors tend to be sparse due to the nature of the entities and relations that are found in large KGs.}. The KG tensor is then factorised to obtain two component tensors, one representing entities and another representing relations \cite{nickel:2016:kg_emb_review}. These tensors capture the latent semantic properties of the entities and the relations respectively. The interaction (via tensor dot product operations) between the entity, relation and entity-transposed tensors provides an indication of the plausability of a relation between two entities.

\begin{figure}
    \centering
    \includegraphics[scale=0.6]{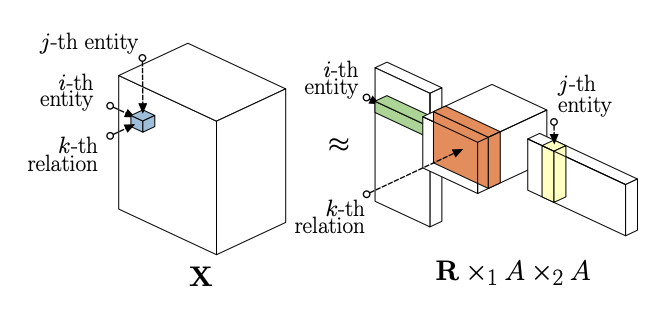}
    \caption{Factorization of an adjacency tensor X (left) using the RESCAL model. The interaction, via dot product, of the factorised A matrix (containing the latent information about entities) and \textbf{R} matrix (in orange on the right) as well as the transposed A matrix, provide information about the relation between pairs of entities. Source: \cite{nickel:2013:rescal}}
    \label{fig:rescal}
\end{figure}

The \texttt{RESCAL} model by Nickel et al \cite{nickel:2013:rescal} is representative of this approach and is depicted graphically in \autoref{fig:rescal} (\autopageref{fig:rescal}). RESCAL's tensor factorisation approach has, however, a higher complexity compared to TransE and TransH\cite{wang_etal:2017:kg_embeddings}. It has a space complexity of \begin{math}\mathcal{O}\end{math}(\textit{nd} + \textit{md}$^2$) and a time complexity of \begin{math}\mathcal{O}\end{math}(\textit{d}$^2$), compared to a space complexity of \begin{math}\mathcal{O}\end{math}(\textit{nd} + \textit{md}) and a time complexity of \begin{math}\mathcal{O}\end{math}(\textit{d}) for TransE and TransH\footnote{Where \textit{n} and \textit{m} denote the number of entities and relations in the KG being embedded, and \textit{d} denote the dimensionality of the entity embedding space.}. 

Another latent feature models, \texttt{DistMult} \cite{yang:2014:distmult} addresses the complexity issue faced by RESCAL by restricting the matrix, that represents relations, to be a diagonal matrix \cite{wang_etal:2017:kg_embeddings}. This reduces the number of parameters for representing each relation to \begin{math}\mathcal{O}\end{math}(\textit{d}). While DistMult addresses the complexity issue faced by RESCAL, and the representation approach taken by both of them \textit{``scale well and can naturally handle both symmetry and (ir-)reflexivity of relations; using an appropriate loss function even enables transitivity''}\cite{trouillon_etal:2016:complex}, both models will require a substantial increase in the dimension of the matrices in order to represent asymmetric relations. 

The \texttt{ComplEx} \cite{trouillon_etal:2016:complex} model proposed by Trouillon et al (2016) extends the expressivity of DistMult by leveraging the properties of complex numbers, where the result of a dot product of two complex-valued matrix is asymmetrical\footnote{\textbf{\texttt{A}} $\cdot$ \textbf{\texttt{B} $\neq$ \textbf{\texttt{B}} $\cdot$ \textbf{\texttt{A}}}, where \textbf{\texttt{A}} and \textbf{\texttt{B}} are complex-valued matrices.}. This property allows ComplEx to represent asymmetric relations with the same level of complexity as DistMult. 

%%%%% I don't know where this section should go 

\subsection{Augmenting word embeddings with knowledge graph embeddings\label{subsec:Augment}}
There are many ways that experimenters have tried to augment embeddings with information encoded from lexical resources. Because many lexical resources are available in the form of an ontology \cite{miller1995wordnet, baker1998berkeley}, a portion of these approaches have been geared towards learning an alignment of word embeddings to knowledge graph embeddings \cite{camacho2018word}.

There are two broad approaches concerning the alignment of ``static'' word embeddings with KG embeddings. Firstly, \textit{\textquote{joint specialization models modify the learning objective of the original distributional model by integrating the constraints into it}}. \cite{fang2016entity} for example, will modify the learning objective in order to learn the distribution of the entity-to-entity co-occurrence (rather than co-occurrence of a word in its textual context), whenever possible. On the other hand, post-processing models will carry out what is commonly called retro-fitting \cite[35]{glavavs2018explicit}. Retro-fitting involves fine-tuning pre-trained word embeddings by presenting constraints. For example, in the case of retro-fitting word embeddings with KB information, such constraints introduced would be similarity constraints between two particular entities. There have been many experiments which take this approach, namely \cite{faruqui2014retrofitting}, who develops a generalized graph-based method to integrate lexical resources. This method iteratively fine-tunes word embeddings, altering the euclidean distance between vectors to better respect related and similar words provided by the WordNet FrameNet and paraphrase database (PPDB) KBs. 

The current existing joint contextualized embeddings and KG embeddings models have been achieved by implementation of an attention mechanism \cite{sun2019ernie, peters_etal:2019:knowledge_enhanced_contextual}. As attention mechanisms and the methods in these experiments are specific to the architecture of BERT embeddings, these techniques could not be applied as such to ELMo's contextualized embeddings. To our knowledge, there are no existing models which could be used to augment ELMo embeddings with knowledge base information. 

%%%%%%%%%%%%%%%%%%%%%%%%%%%%%%%%%%%%%%%%
\subsection{Evaluating embedding models}
%\todo{change knowledge graph into KB wherever necessary}
There exists two main approaches to embedding model evaluation. The first family of methods relies on tasks that we try to achieve using the embeddings and then score the performance on the tasks.
With the evaluation being done not at the level of the embedding model but later in the processing, such methods are usually called \emph{extrinsic evaluation methods}.
They are sometimes referred to as \emph{downstream tasks} as such a task is used to evaluate the model.

In contrast, the second family of methods does not rely on external tasks and are thus called \emph{intrinsic evaluation methods}.
They typically rely on metrics applied directly to the embeddings or the embedding space, so at the level of the embedding model.

% difficulty of data type for extrinsic, evaluated data for intrinsic

Both intrinsic and extrinsic evaluations of an embedding model can be interpreted in terms of linguistic properties of language the model should capture (like similarity and relatedness, see \cite{salahli:2009:relatedness}).
One should note however that it is only an interpretation of the results and in the case of downstream tasks, of the way the task is designed.

\subsubsection{Intrinsic evaluation}
According to \cite{bakarov:2018:review_word_embeddings}, \textit{\blockquote{methods of intrinsic evaluation are experiments in which word embeddings are compared with human judgments on words relations}} (in the case of word embeddings).
They categorize intrinsic evaluation methods using human judgments in 2 major categories:
\begin{enumerate}
    \item absolute evaluation, where automatic metrics applied on human annotated data;
    \item comparative evaluation, in which humans evaluate the model by comparing the outputs of multiple models, such method typically relying on crowd-sourcing.
\end{enumerate}
They further categorize those methods based on how the human assessment is collected, namely conscious evaluation (with time to think about the decisions) and unconscious evaluation (without time to think about the decisions or with the evaluation being peripheral to what the evaluator is asked to do).

They mention that this typology is not exhaustive as it does not account for methods that do not rely on human judgment, like those comparing the results of textual embedding to a reference KB or those based on intrinsic properties of language. In a more general view, this last category includes the methods using the structure of the data training data (text or KB) to evaluate the embedding models.

Here is a list of the main intrinsic evaluation methods present in \cite{bakarov:2018:review_word_embeddings,gladkova_drozd:2016:intrinsic,torabi_etal:2018:word_embeddings}.
\begin{description}
    \item[Semantic similarity-based methods] assume that the distance of the elements in the embedding space represents their similarity.
        Closer elements would be more similar.
        Typically, the distance measure used is the cosine distance.
        A wide range of datasets of pairs of words with human similarity scores is available, as it is the most widespread intrinsic evaluation method for embeddings.
        This method is a straightforward evaluation of the modeling of similarity.
        However as explained in \cite{gladkova_drozd:2016:intrinsic} this kind of method is biased by what the annotators understand by semantic similarity.
        This method is a straightforward evaluation of the modeling of similarity.
    \item[Synonymy-based methods] are based on the detection from a set of elements the closest to a given element. This kind of method is very close to semantic similarity-based methods as it is based on the same assumptions and can be interpreted in the same way. Synonymy-based methods are more clearly linked to similarity however, as they explicitly evaluate synonymy.
        
%    \todo{finish the description}
    % ""is based on the idea that word similarity could be measured through ﬁnding the most similar word relative to a set of other words""
    \item[Analogy-based methods] are based on the idea that if the embeddings correctly encode the attributes of the elements encoded, we should be able to execute basic arithmetic using the embeddings.
        A stereotypical example is that of \emph{king - man + woman = queen}, with \emph{queen} being to \emph{woman} what \emph{king} is to \emph{man}.
        A major drawback of this family of methods is the lack of precise evaluation metric.
        We can consider this method as an evaluation of the ability to model relatedness.
        %\todo{finish the description past this point}
    \item[Thematic fit-based methods] evaluate the ability to determine semantic roles of a predicate, in other words, the ability of the model to propose the most relevant noun to use with a given verb and a given role.
        It can be seen as a measure of the similarity or of the ability of the model to learn to categorize the elements by their semantic role.
        
    \item[Semantic (embedding) space coherence-based methods] try to evaluate the quality of the neighborhood of elements in the embedding space.
        This can be interpreted as an evaluation of the internal coherence of the embedding space.
        As described in \cite{gladkova_drozd:2016:intrinsic}, a typical task to propose to the human evaluator would be to select an element and its 2 nearest neighbors in the embedding space as well as a randomly chosen element.
        The evaluator should be able to determine that the randomly chosen one is an outlier, an \textquote{anomaly}.
        %\cite{gladkova_drozd:2016:intrinsic}, https://trello.com/c/9u2c1VqO
        
    \item[Linguistic features alignment methods] described in \cite{gladkova_drozd:2016:intrinsic} evaluate how well the dimensions of a word embedding can be aligned with dimensions of so-called linguistic vectors computed from a semantically annotated corpus.
        Because we rely on a semantically annotated corpus we do not need direct human annotation or rating for this task.
        The results of this evaluation method have been shown to be fairly correlated to performance in downstream tasks, making it a good performance estimate.
        Also, by comparing the embedding with linguistically justified dimensions, this method focuses on how easily we can interpret the embedding themselves according to linguistically meaningful dimensions.
        %\cite{gladkova_drozd:2016:intrinsic},  https://trello.com/c/9u2c1VqO
        
    \item[Clustering methods] are methods for which \blockquote[\cite{bakarov:2018:review_word_embeddings}]{the task is to split [a set of words] into subsets of words belonging to different categories} and the evaluation of the quality of the clustering by comparing it to the known categorisation.
        \blockquote[\cite{bakarov:2018:review_word_embeddings}]{Possible criticism of such method could address the question of either choosing the most appropriate clustering algorithm or choosing the most adequate metric for evaluating clustering quality.}
        Also, in our opinion, such methods are more extrinsic than intrinsic because they rely on a additional model to solve a task.
        Indeed, from the own words of \cite{bakarov:2018:review_word_embeddings}, clustering evaluation methods are based on a task.
        Such method can be interpreted as a measure of whether the model learns to categorize the elements in a similar manner as the reference categorization.
        %
        % Given a set of words, the task is to split it into subsets of words belonging to diﬀerent categories (for example, for words dog, elephant, robin, crow the ﬁrst two make one cluster which is mammals and the last two form another second cluster which is birds; the cluster name is not necessary to be formulated) [Baroni et al., 2014]. The amount of clusters should be deﬁned. Possible critique of such method could address the question of either choosing the most appropriate clustering algorithm or choosing the most adequate metric for evaluating clustering quality.
    \item[Outlier detection methods] also rely on a task which is to identify the element which is different from the rest within a provided set of elements, based on a known categorisation.
        Those methods are quite similar to clustering methods in that they deal with element categories, and can thus be interpreted in a similar manner and are, in our opinion, more extrinsic than intrinsic methods.
        %""evaluates the same feature of word embeddings as the Clustering method, but the task is not to divide a set of words into certain amount of clusters, but to identify a semantically anomalous word in an already formed cluster (for example, for a set {orange,banana,lemon,book,orange} which are mostly fruits, the word book is the outlier since it is not a fruit)""
\end{description}

%A usual method the metric tosimilarity w/ human scoring using Spearmann's rank correlation coef

%\cite{gladkova_drozd:2016:intrinsic}
%\cite{torabi_etal:2018:word_embeddings}
% overview of existing intrinseque [2, 3 word & KB]
%   for ctx [1, 5]
%   for KB [4 defines ProNorm]

%alse performance of the embedding on training/evaluation data, in the case of autoencoder 

As mentioned in \cite{bakarov:2018:review_word_embeddings}, \textit{\blockquote{NLP engineers who are more interested in dealing with downstream tasks (for instance, semantic role labeling) usually evaluate the performance of embeddings on such tasks, while computational linguists exploring the nature of semantics end to investigate word embeddings through experimental methods from cognitive sciences.}}
This insight can serve as an hint to why embeddings-related literature focuses on downstream tasks, even if intrinsic evaluation allows for a more targeted analysis of the inner workings of an embedding model.
It is indeed a field where the practical use of an embedding model is a core factor of whether the model will be used or not.

\subsubsection{Extrinsic evaluation}
In practice, extrinsic evaluation of embedding models involves a task or set of tasks, the choice of which should be guided by the properties of the embedding model that we want to evaluate.
Such a task is usually defined by:
\begin{enumerate}
    \item the goal of the task (ex: detecting similar words, predicting relations, \textit{etc.});
    \item a machine learning model separate from the embedding models, that will use the embeddings as features to solve the task;
    as mentioned by \cite{bakarov:2018:review_word_embeddings}, such a model is typically supervised;
    \item data on which the embedding models will be applied to obtain the embeddings and perform the task;
    \item a way to evaluate the performance on the task.
\end{enumerate}

A wide variety of downstream tasks have been designed over the years to evaluate embedding models, the main ones described in \cite{bakarov:2018:review_word_embeddings} and \cite{wang_etal:2017:kg_embeddings} respectively for word and KB embedding models. Most methods used for word embedding models are applicable as-is on contextual embeddings, though with slightly different data (\cite{torabi-asr-etal-2018-querying}).

While the tasks are quite similar between textual (word or context) and KB embedding model evaluations, the kind of data used as input for the embedding model is quite different.
For textual embeddings the data is text, while for KB embeddings the data is either triples or KG node identifiers.

We will describe the main families of tasks used and how they are designed for both textual and KB embeddings. For an overview of the datasets available for each task, see the corresponding sections in either \cite{bakarov:2018:review_word_embeddings} or \cite{wang_etal:2017:kg_embeddings}.
\begin{description}
    \item[Relation (or link) prediction] is a type of task focused on identifying the relationship between two elements.
        While very used for KB embeddings, \cite{peters_etal:2019:knowledge_enhanced_contextual} presents a contextual embedding version.
        
        To perform this task with KB embeddings, the data used is a triple \texttt{<subject, \_, object>} in which the relation has been removed and serves as the target of the prediction task.
        For textual embeddings, we have something of a textual adaptation of the KB version of the task, with the object and subject of the relation lexicalized --~put into a textual form, either a word or a word in context.
        There is a variant of textual relation prediction called \emph{relation extraction}, in which the relation to predict is described in the text (ex: \textquote{Barrack Obama was elected president of the USA}, the relation of being the president of is present in the text) making the task more of an extraction of the relation from the text than a simple prediction of the relation between entities.
        
        While usually what is predicted is the relation between the elements, the relation can be provided with the source or the target of the relation, the remaining one being the prediction target. In that case, it can be called \emph{entity prediction}.
        Relation and entity prediction are more generally called \emph{triple completion} as they amount to filling the missing part of a triple.
    \item[Entity typing (or classification)] is the task of predicting the type of an entity represented by a KG node or a word.
        Entity typing is typically used for KB embeddings as type information is usually included in the knowledge base, but \cite{peters_etal:2019:knowledge_enhanced_contextual} presents a contextual embeddings version.
        This family of tasks can be interpreted as a measure of whether the model learns to categorize the elements by their nature.
        
        With KB embeddings we embed a single entity and try to determine its type.
        It is usually quite easy to acquire the gold labels for this task with KG, as typically every node is attributed a class.
        With textual embeddings we take a word (and its context if necessary) and predict its type in a similar manner.
        It is however a bit less straightforward to determine the type of the entity in this case than with KBs.
    \item[Entity resolution] is the task of checking if two words or KG nodes represent the same entity.
        This kind of task is supposed to evaluate the ability to model semantic similarity in a rather straightforward manner.
        For textual embeddings in particular, the task can be used to evaluate how synonymy is encoded in the model.
        
        For KB embeddings we embed pairs KG nodes known to represent or not the same object, while in the case of word embeddings, we embed pairs of words.
        
        There is also variant of the task called \emph{word in context} (WiC) \cite{peters_etal:2019:knowledge_enhanced_contextual} which corresponds to the contextual embedding version of the task, for which we determine if words used in their respective contexts are used with the same sense.
    
    \item[Textual entailment detection] is a task for which we determine if a chunk of text logically entails another.
        A typical example would be \blockquote{Mary gave John a book} that implies that \blockquote{Mary gave a book}.
        This task should evaluate the ability of the model to encode logical relations, and fits more contextual embeddings and so-called sentence embeddings than word embeddings.
    \item[Triple (fact) classification] is a task for which we determine if a triple describes a true information or not.
        Reusing the example from \cite{wang_etal:2017:kg_embeddings}, the triple <Alfred Hitchcock, Director Of, Psycho> is true while <James Cameron, Director Of, Psycho> is false.
\end{description}

Those generic task families are the most used because they have a wide variety of datasets available and are quite simple to set up.
Also, while those relatively simple tasks do not evaluate performance for a target application, they can be used to estimate the performance when we rely on properties of the language similar to those evaluated by the task.
Such an approach was used in \cite{peters_etal:2019:knowledge_enhanced_contextual}, where the authors use relation extraction, word in context and entity typing to evaluate their knowledge enhanced contextual embedding.
While their method is close to what we propose, they evaluate their model exclusively on textual data.
    
Tasks which are more closely related to real world use cases are also used, for example  triple extraction from text,
question answering, recommender systems, Named Entity Recognition (NER) or semantic role labelling.
Those tasks are harder to analyse to linguistically interpret the performance of the model, but they provide a better insight on how the embedding model impacts performance for a target application.

% overview of existing extrinsic
%   for ctx [1, 5]
%   for KB 

%%%%%%%%%%%%%%%%%%
%This process allows to
%metrics oriented
%   pros & cons of each [1]

%   works comparing [1, 2, ]

% intrinseque VS extrinsique
%   reminder of the terms
%       intrinseque
%       extrinsique
%   pros & cons of each [1]
%   works comparing [1, 2, ]
% overview of existing intrinseque [2, 3]
%   for ctx [1, 5]
%   for KB
% overview of existing extrinsique []
%   for ctx [1, 5]
%   for KB [7!!, 4, 8 clustering & link]

% articles
%1  bakarov:2018:review_word_embeddings
%2  gladkova_drozd:2016:intrinsic
%       best model closest to human evaluation
%3  simov:2017:extrinseque_common
%4  jouravlev:2016:relatedness
%5  torabi_etal:2018:word_embeddings
%6  perone_etal:2018:sentence_embedding
%7  wang_etal:2019:word_embeddings
%7  wang_etal:2017:kg_embeddings
%8  goyal_etal:2018:graph_embedding_survey

% what we do is very close to what is done in peters_etal:2019:knowledge_enhanced_contextual (kb-extended contextual), but extebded to kb
\section{Approach\label{sec:approach}}
In the following section we develop the approach taken in establishing a framework of comparative analysis between the use of contextualized embeddings and KG embeddings as input of a Machine Learning NLP problem. We first define the two tasks that we have 
% add a sentence to say we have 4 subsections: hypothesiis + 2 subtasks + chosen pretrained models

\iffalse
\subsection{Methods}
This survey provides a two-steps contribution:
(1) we propose a methodology to build new datasets for our both tasks. We extract the context of each entity from entity descriptions corresponding to paragraphs of text. The context is only one sentence containing the entity and is selected according to two approaches. The selected sentence must be the closest to the lexicalization. To operate this selection, we first compute the proportion of tokens of the lexicalization with a TF-IDF measure in the sentence. Secondly, as lexicalization can contain spaces, we set a maximum limit distance between tokens and minimize the best distance.
For each task, our model takes two different types of entities as inputs.
The first input is the word embedding obtained by giving to ELMo the context in which the word is located.
The second input is the KB embedding identity obtained with the training of the KB embeddings.
%to be completed for the (2)
(2) In addition, the first task tries to predict the entity type as described in 3.3.
The second task repeats the input two times because it tries to predict the relation between two entities as explained in 3.4.
\fi
% les deux genres de données en parallèle: text + triplets
\subsection{Task 1: entity typing\label{subsec:approach-task1}}
%\blockquote[\cite{choi2018ultra}]{Incorporating fine-grained entity types has im-proved entity-focused downstream tasks, such asrelation extraction (Yaghoobzadeh et al., 2017a), question  answering  (Yavuz  et  al.,  2016),  queryanalysis (Balog and Neumayer, 2012), and coref-erence resolution (Durrett and Klein, 2014). . . We observe that text often contains cuesthat explicitly match a mention to its type, in theform of the mention’s head word.   For example,“the  incumbent  chairman  of  the  African  Union”is  a  type  of  “chairman.”This  signal  complements the supervision derived from linking entitiesto  knowledge  bases,  which  is  context-oblivious.}
Given a text with the label available for the entity and the relation between them, the aim of the entity type prediction is to detect and predict semantic types of a named entity in knowledge graphs. As entities can have multiple types, we are predicting the set of types an entity should have.
It is beneficial for a large number of NLP tasks such as entity linking, relation extraction, question answering and knowledge base population.
Linking this task to our previous hypothesis, entity typing does not directly evaluate similarity nor relatedness, but something like meaningfulness of the representation.

\subsection{Task 2: relation prediction\label{subsec:approach-task2}}
Relation prediction is a task that take two entities and predict the relations between these two elements. We have the sentences that contains information on the labels of the entities, their location in the sentence, but not the relation between the entities. As this task is directly linked with the concept of relatedness, it allows us to verify our previous hypothesis.

%\todo{add the respective size of the embeddings of elmo (512) \& biggraph (400)}

\subsection{Chosen frameworks\label{subsec:approach-chosen}}
%add why we choose to use pretraind models exclusively

We chose to conduct our experiments with ELMo. It has a bidirectional language modelling architecture that is based on a predict neighbouring-word task. Compared to other contextual embedding models such as BErT\footnote{BErT utilises multi-head attention, and its training tasks include token masking and next sentence prediction.}, ELMo has a relatively less complex architecture and its training task is the same as that used in classical \textquote{static} embedding models such as Word2Vec, albeit ELMo is trained on the next as well as the previous world. The neighbouring word language modelling task is well-studied, and allowed us to better explain the results of our experiments. We used the large version of the official ELMo pre-trained model\footnote{\url{https://allennlp.org/elmo}}, where each contextualised word is represented by a 512-dimension vector.
%\todo{to fact check, discuss choice as a group and possibly refine}

For KG embeddings, we chose to work with ComplEx because it has been shown to perform well on KGs. On the FB15k dataset, ComplEx was shown to perform better %\todo{I don't get what task they tested the different models with FB15K, \url{http://proceedings.mlr.press/v48/trouillon16.pdf}, can someone help take a look too?}
compared to TransE and DistMult on the prediction of relations in the FB15k dataset (see \autoref{tab:kg_emb_comparison}, \autopageref{tab:kg_emb_comparison}). To obtain the ComplEx embeddings, we used the PyTorch-BigGraph\footnote{\url{https://github.com/facebookresearch/PyTorch-BigGraph}} implementation of ComplEx \cite{lerer:2019:pytorch_BG}, trained on FB15k (see \autoref{subsec:data-fb15k} \autopageref{subsec:data-fb15k}), a subset of Freebase. The resulting ComplEx embedding model represents entities as 400-dimension vectors.

\begin{table}
    \centering
    \begin{tabular}{lll}
    \toprule
    \textbf{Model} & \textbf{MRR} & \textbf{Hits@10} \\\midrule
    TransE & 0.221 & 0.641 \\ 
    DistMult & 0.242 & 0.824 \\ 
    ComplEx & 0.242 & 0.840 \\\bottomrule
    \end{tabular}\\[0.5em]
    \caption{Comparison of the performance of various KG Embedding approaches on the FB15k dataset. Source: \cite{trouillon_etal:2016:complex}}
    \label{tab:kg_emb_comparison}
\end{table}

\subsection{Hypothesis\label{subsec:hypothesis}}
Starting with the assumptions that KB embeddings represent entities in triples and word embeddings represent entities from unlabeled and unstructured text, our initial hypothesis is that KB embeddings are most likely to obtain significant scores in entity typing and relation prediction, when compared to with word embeddings. According to the structure of the triples, entities are easier to spot in KB embeddings.
Our second hypothesis should rely on the fact that a combined model that join the KB embeddings and the contextual embeddings can perform at least equally or better at most than the maximum possible score for one model on one task.
Another point to be mentioned is the similarity dimension where contextual embeddings are supposed to perform better than KB embeddings. This point will not be exploited in the next steps of our work.
\section{Data\label{sec:data}}

% \todo{what kind of data we need
% corpus used
% creation methods
% description of dataset produced
% [Phlee + Kelvin]}

The overall aims of our experiments being to compare the efficacy of contextualized embeddings with knowledge graph embeddings, this factored into the kind of data required for experiments. As previously mentioned, in order to access ELMo embeddings of a given word, we are required to feed a full sentence in which it is situated. For this particular reason, both experiments necessitated not only the lexicalization of the entities themselves, but of these entities in their context. On one hand, for the task of entity typing, we required a sentence in which situates each lexicalized entity. On the other hand, for the task of relation prediction, we would need a dataset containing lexicalizations of triples, not just the entities themselves. In order to carry out experiments for both of our tasks with consistency, we sought to use datasets and resources which are based on the same KB. We found the Freebase KB to be the most appropriate to fit these needs. 

Freebase is an open collaborative KB launched in 2007 by Metadata and bought by Google in 2010. The project was brought to a close, as Google migrated the data to Wikidata in 2015 in order to support this fast growing effort to lead an open collaborative KB \cite{pellissier2016freebase}. The motivation to basing our experiments off of this KB are as follows:
\begin{itemize}
    \item an existing subset of this KB, Freebase 15K \cite{bordes_etal:2013:transE}, makes experiments more manageable given our computational power constraints;
    \item there are existing extensions of the Freebase 15K which are appropriate for the typing task   \cite{xie2016representation};
    \item there are available lexicalizations of Freebase triples based off distant supervision \cite{riedel:2010:fbnyt}.
\end{itemize}

\subsection{Freebase15K\label{subsec:data-fb15k}}

Freebase 15K (FB15K) is a subset of the full Freebase KB, which has been extracted and pre-processed by \cite{bordes_etal:2013:transE} in order to carry out preliminary experiments. In order to produce the subset FB15K, experimenters targeted entities and relations which are also present i the still existing Wikidata \cite{vrandevcic2014wikidata} database while maintaining at least 100 mentions in the larger Freebase. It has been widely used for experiments in both embedding knowledge graph embeddings \cite{wang_etal:2014:transH, lin:2015:transR, xie2016representation} as well as in entity-focused tasks such as entity typing \cite{xin2018improving}.

The FB15K dataset contains entity codes, the lexicalisations of these entities, their types, relations, and the triples split into train, valid, and test datasets. Counts for the entities and relations present in the FB15K  can be seen in \autoref{tab:fb15kcounts} (\autopageref{tab:fb15kcounts}). 

\begin{table}[h]
    \centering
    \begin{tabular}{ll}
    \toprule
    & Count \\\midrule
         Entities & 14,951 \\
         Relationships & 1345 \\
         Types & 4054 \\
         Train set & 483,142 \\
         Valid set & 50,000 \\
         Test set & 59,071 \\\bottomrule
    \end{tabular}\\[0.5em]
    \caption{Elements and triple counts in FB15K}
    \label{tab:fb15kcounts}
\end{table}

\begin{table}[h]
    \centering
    \begin{tabular}{ll}
    \toprule
    & Count \\\midrule
       Named entities  & 13533 \\
       Non named entities & 1408 \\\bottomrule
    \end{tabular}\\[0.5em]
    \caption{Counts of named entities in FB15K}
    \label{tab:fb15ktypes}
\end{table}

\begin{table}[h]
    \centering
    \begin{tabular}{ll}
    \toprule
    & Count \\\midrule
        Train & 11607 \\
        Test & 1268\\\bottomrule
    \end{tabular}\\[0.5em]
    \caption{FB15K sample counts after filtering}
    \label{tab:fb15kpost_filt}
\end{table}

\begin{figure}
    \centering
    \includegraphics[width=\textwidth]{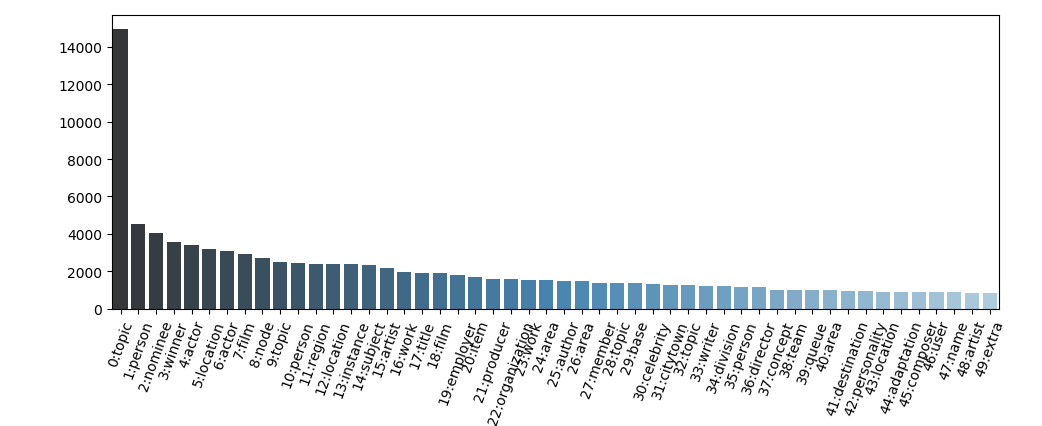}
    \caption{Distribution of entity types}
    \label{fig:my_label}
\end{figure}
%include tables and/or graphs of distribution of dataset. Include: 1. general count of # of each entity type. 2. if possible, multi-class distribution 

As previously mentioned, we not only needed a manageable sized KB to work from for our tasks, but we also necessitated on one hand, sentences in which a given entity is situated, and on the other, surface realizations of triples (discussed in \autoref{subsec:data-fbnyt}). For this reason, along with the FB15K dataset, we have made use of an extension of this dataset containing descriptions of the entities which was extracted by experimenters \cite{xie2016representation}. In order to process this dataset, we followed the following procedure: 
\begin{enumerate}
    \item look up and tokenize the full lexicalization of the entity code;
    \item select entity head-word based on most frequent token across description;
    \item search for mention in its corresponding description;
    \item drop items whose mentions do not match the lexicalization (with allowance of up to 1 token, as in a middle name, which separates original tokenized lexicalization);
    \item select candidate sentences with mentions matching this criteria;
    \item take first sentence from candidate list, using headword and it's index in the sentence as position in context.
\end{enumerate}

Examples \ref{ex1} and \ref{ex2} show instances where an entity description did not pass through our pre-processing would be.
\begin{exe}
\item \textsc{Actor-GB:} \label{ex1}An actor is a person portraying a character in a dramatic or comic production; she or he performs in: film, television, theatre, or radio \emph{etc.}
\item \textsc{Emily Blunt:} \label{ex2}Emily Olivia Leah Blunt is an English actress. She has appeared in The Devil Wears Prada, The Young Victoria, The Adjustment Bureau, and Looper. She has been nominated  \emph{etc.}
\end{exe}

The previous examples did not carry over for the reasons that the entity mention (appearance in the text) did not match that of our input data, or there were more than one tokens interrupting the succession of tokens which compose the entity lexicalization. In example \ref{ex3} our processing took hold of the entity. In the example the selected head-token is underlined, the full lexicalization in bold, and the context sentence in italics.
\begin{exe}
\item \textsc{E1 Music:} \label{ex3}\textit{\textbf{E1 \underline{Music}}, the primary subsidiary of Entertainment One LP, is an independent record label in the United States.} It is widely regarded as the most successful independent record label in the United States, having garnered the most Billboard hits of any independently-owned music label in history \emph{etc.} 
\end{exe}

\subsection{Freebase-NewYorkTimes\label{subsec:data-fbnyt}}

The Freebase-NewYorkTimes dataset (``FB-NYT'') contains surface realisations of entities and relations found in Freebase \cite{riedel:2010:fbnyt}. The surface realisations are from the New York Times annotated corpus\footnote{\url{https://catalog.ldc.upenn.edu/LDC2008T19}} released by the Linguistics Data Consortium, which is made up of 1.8 million news articles that were published by the American newspaper, The New York Times, in a 20-year period between 1987 and 2007. 

The dataset was built automatically\footnote{This was done with the following steps (i) named entity recognition, on all of the sentences in the New York Times annotated corpus, (ii) string match of the entities mentions found with the canonical form of the entity name in Freebase, (iii) feature extraction of the entities matched; and (iv) factor graph and constraint-driven semi-supervision to jointly decide if a relation exists between two entities as well as what the relation is.} on the following basis\footnote{This is a relaxed version of the original \textit{distant supervision assumption} that \blockquote{if two entities participate in a relation, all sentences that mention these two entities express that relation.}}: 
\blockquote[\cite{riedel:2010:fbnyt}]{If two entities participate in a relation [as part of a triple in a KG], at least one sentence [in a text corpora] that mentions these two entities might express that relation.} Although the dataset was automatically created, manual evaluation of a sampling of the generated FB-NYT with two human annotators indicates, with statistical significance, that the approach is able to identify relation instances with a precision of 91\%. 

FB-NYT is suitable for our relation type prediction task as it contains the surface realisation of triples found in Freebase. The surface realisation provides us with the context that allows us to obtain the contextual word embeddings for the entities. However, our experiments are conducted with knowledge graph embeddings trained on a subset of Freebase, FB15k. As such, we filtered FB-NYT for relations and entities that were found in FB15k. \autoref{tab:fbnyt-stats} (\autopageref{tab:fbnyt-stats}) provides an overview of the sentences, entities, relations and samples before and after our filtering.

\begin{table}
    \centering
    \begin{tabular}{lll}
        \toprule
         & Before filtering & After filtering \\\midrule
        Number of sentences & 66,196 & 13,874 \\
        Number of entities & 14,664 & 798 \\
        Number of relations & 24 & 16 \\
        Number of samples & 111,327 & 29,492 \\\bottomrule
    \end{tabular}\\[0.5em]
    \caption{Composition of FB-NYT dataset before and after our filtering}
    \label{tab:fbnyt-stats}
\end{table}

\begin{figure}[h]
    \centering
    \includegraphics[scale=0.5]{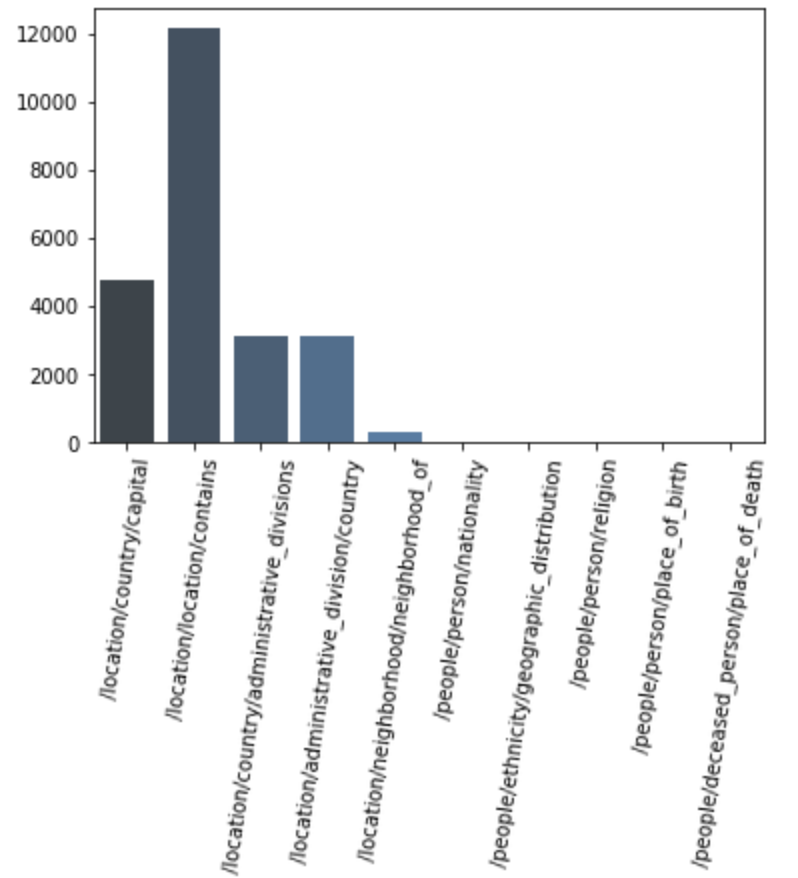}
    \caption{Distribution of relation types in the training set of our filtered FB-NYT corpus}
    \label{fig:relation_types}
\end{figure}

% \section{Datasets}
% adapting existing corpora to apply the tasks with both KB & contextual embeddings

% \subsection{Existing datasets used}
% % describe "source corpus" used
% %   corpus 1
% %       description of the corpus & where to find it
% %       structure of the data
% %   corpus 2
% %       description of the corpus & where to find it
% %       structure of the data
% %   corpus 3
% %       description of the corpus & where to find it
% %       structure of the data
% %   ...

% \subsection{New datasets created}
% %describe corpus we created
%   for task typing
%       structure of the data we need for the task to be usable equally for KB & contextual embeddings
%       from which "source corpus"
%       aggregation/creation process
%       data/label example(s)
%       statistics (number of samples, kind of entities, number of types, kind of types, ...)
%   for task relation
%       structure of the data we need for the task to be usable equally for KB & contextual embeddings
%       from which "source corpus"
%       aggregation/creation process
%       data/label example(s)
%       statistics (number of samples, kind of entities, number of relations, kind of relations, ...)

\section{Experiments\label{sec:experiments}}
We present in this section the main aspects of our experimental setup: how do we explore the possibility of combining knowledge and text in the embedding, how do we perform the tasks and how we determine the performance of the models on the tasks.

\subsection{Proof-of-concept combined model}
For our experiments we introduce a simple embedding model combining KB and contextual information in the same embedding.

% HOW/WHAT:
%   simply concatenate the embeddings obtained with each kind of data
%   result is a fairly large embedding containing at the same time the information from two representations of the same thing
The model simply concatenates the embedding generated with the two pre-trained models we use for KB and contextual data (BigGraph and ELMo respectively).
The resulting embedding is almost twice as large as each of the separate embeddings (512 cells from ELMo and 400 from BigGraph so 912 cells once concatenated).
However, it contains the information of both the KG node representation and the textual description exactly in the same way as ELMo and BigGraph do separately, without removing or adding anything.
As such, this model is likely to be only at the surface of what can be gained by embedding text and knowledge together.

% WHY:
%   check whether using KB & contextual data on the same level works
%   try to improve the performance (in other words, check with a straightforward method if adding KB or contextual representation improve the performance significantly)
%       potential implications if performance is improved (may be removed from the report): encourage to enrich one kind of data if necessary, serve as a baseline for more complex models
%       if performance is not improved: no conclusion can be drawn in that regard, because the model is most likely too simple or the classifier not complex enough to handle the data
This model is meant as a proof-of-concept to study whether embedding KB and contextual information at the same level can benefit the downstream tasks. Also, if such a model make a significant improvement of the performance relative to a KB or contextual representation alone, this result would encourage the enrichment of KB and textual datasets with textual and KB data respectively. Lastly, as the model we use is extremely simple, it can serve as a baseline for more complex models relying on both kinds of data.

\subsection{Task setup}
The tasks we have chosen to develop and evaluate -- namely entity typing and and the relation prediction tasks (using the corpora described respectively in \autoref{subsec:data-fb15k}, \autopageref{subsec:data-fb15k} and \autoref{subsec:data-fbnyt}, \autopageref{subsec:data-fbnyt}) -- are both multi-class classification tasks.
Indeed, for typing we classify an entity embedding into a set of types, and for relation prediction we classify a pair of entity embeddings into the relation between the entities.
A slight difference between the tasks is that we have only one relation to predict per sample (multi-label single-output setting) while we have a varying number of types per sample (multi-label multi-output setting).
We thus use a very similar approach in both cases:\begin{enumerate}
    \item we generate the embedding(s) for the entity or (entities) to classify;
    \item we train a Logistic Regression classifier on the embeddings from the training set;
    \item we evaluate the quality of the prediction using the metrics defined in \autoref{subsec:metrics} (\autopageref{subsec:metrics}).
\end{enumerate}

Our intent with the tasks is not to obtain a high performance on the task but to see how the performance is impacted by the embedding model.
A simple classifier like the Logistic Regression allows us to shift the performance more on the quality of the embedding.
The logistic model may, however, not be complex enough to handle the large dimensionality of the embeddings.

Also, to maintain an \textquote{all other things being equal} setting when building the task, by providing the task module with the same \textquote{blind} embedding model structure taking both the KB and contextual information as input whether the model uses the information to produce the embedding or not.
This allows us to have a systematic classification process, with a single code applied on the embeddings regardless of their input and of the embedding size.

% classification tasks in both cases
% use of logistic regression because we want to evaluate the quality of the model and not the classification model itself
% [implementation-wise]
%   "blind" model structure taking both KB & context + position info to produce the embedding from KB only, ctx only, or both, using strictly the same testing framework
%   systematic classification, with a single code applied on the embedding models with no regard to what kind of data they truly require

\subsection{Evaluation metrics\label{subsec:metrics}}
For our experimental tasks, we used \texttt{Precision@n}, Mean Average Precision@k (\texttt{MAP@k}), and Mean Reciprocal Rank (\texttt{MRR}) as evaluation metrics. These are commonly used evaluation metrics in the field of information retrieval, of which KBs are widely used in. These metrics are relevant when there are multiple labels/answers for a given sample. 

Precision@n is a measure used for data with samples that contain more than one correct label, i.e. a set of labels. It measures, for a single predicted set of labels, the number of these that are accurate up to a set cut-off position, n. It can be expressed in the following form: 

\[\text{Precision@n} = \frac{\text{\# of accurate predictions up to n}}{\text{n}}\]

The MAP@k for a dataset is the mean of the average precision@k (AP@k) for each sample\footnote{For a multi-label single output classification task, MAP@k=1 is the same as simple.}. AP@k is computed as follows: 

\[\text{AP@k} = \frac{\text{1}}{\text{m}} \sum_{i=1}^k \text{P}  (label_k)\text{, if the $k^{th}$ item is relevant}\]

where k is a set cut-off position and m is the number of labels for the particular sample.

The MRR for a dataset is the mean of the reciprocal rank (RR) of each sample in the dataset. MRR and RR are expressed in the following form: 

\[\text{RR} = \frac{\text{1}}{\text{R}}, \quad\quad \text{MRR} = \frac{\text{1}}{\text{N}} \sum_{i=1}^N \text{RR}(output_i)\]

where R is the rank of the accurately predicted label and N is the number of samples in the dataset.

\section{Results\label{sec:results}}

%%%%%%%%%%%%%  task 1  %%%%%%%%%%%%%
For the first experiment, the MAP@k=10 and Precision@k=10 scores in \autoref{tab:results_entity_typing} (\autopageref{tab:results_entity_typing}) suggest that KG embeddings perform better than contextual word embeddings in an entity typing task. This is in line with our starting hypothesis (see \autoref{subsec:hypothesis} \autopageref{subsec:hypothesis}). The results also suggest that a representation, that is a concatenation of both models KG and contextual word embedding models, is almost on par with that of KG embeddings. 

% compare Precision@10 (top 10, but order of the predictions within the top 10 not taken into account) --> Concat [slightly (not significantly) lower] than KB
% compare MAP@10 (top 10, and order of the predictions within the top 10 taken into account) --> Concat [slightly (not significantly) higher] than KB
% even if not significant, as MAP better evaluates the quality of the prediction, Concat barely better than KB

%%%%%%%%%%%%%  task 2  %%%%%%%%%%%%%
For the second experiment, in \autoref{tab:results_relation_prediction} (\autopageref{tab:results_relation_prediction}), suggests that KG embeddings perform better than contextual word embeddings in the task of relation typing. Again, this is in line with our starting hypothesis (see \autoref{subsec:hypothesis} \autopageref{subsec:hypothesis}). Contrary to our initial hypothesis however, the performance of the concatenated representation performs poorer than contextual word embeddings in the same task -- the MRR for the former is 1.2 percentage points below that of the latter. One factor for the poorer result of the concatenated representation is that it may carry duplicated information that is found in both the KG and contextual word embeddings. Further work and analysis \autoref{sec:futurework} (\autopageref{sec:futurework}) would be required to validate this hypothesis.  

% MRR is clearer than Precision (perfect prediction rate) in terms of interpretability of the results, but they complete each other 

\begin{table}[h]
    \centering
    \begin{tabular}{lll}
    \toprule
    \textbf{Model}              & \textbf{MAP@k=10}  & \textbf{Precision@k=10 (mean \textpm{} std)}\\
    \midrule
    Contextual Embeddings (1)   & 0.631 & 0.449 \textpm{} 0.271 \\
    KG Embeddings (2)           & \textbf{0.825} & 0.528 \textpm{} 0.269 \\ 
    Concatenation (1) + (2)     & \textbf{0.828} & 0.527 \textpm{} 0.268 \\
    \bottomrule
    \end{tabular}
    \caption{Results of the entity typing experiment}
    \label{tab:results_entity_typing}
\end{table}

%%%%%%%%%%%%%%%%%%%%%%%%%%%

\begin{table}[h]
    \centering
    \begin{tabular}{lll}
    \toprule
    \textbf{Model}              & \textbf{MRR}  & \textbf{Precision (MAP@k=1)}\\
    \midrule
    Contextual Embeddings (1)   & 0.750   & 0.554 \\
    KG Embeddings (2)           & \textbf{0.817}   & 0.663 \\
    Concatenation (1) + (2)     & 0.738   & 0.533 \\
    \bottomrule
    \end{tabular}\\[0.5em]
    \caption{Results of the relation prediction experiment}
    \label{tab:results_relation_prediction}
\end{table}

%%%%%%% EXTRA STUFF %%%%%%%% 

%%% For Task 1
% We note however that there is a significant difference between the precision as well as the MRR scores of the Contextual Embeddings and KG Embeddings. 
%%% For Task 2
%We note that there is a significant class imbalance in our relation typing task (See \autoref{fig:relation_types} (\autopageref{fig:relation_types})). A common approach to address class imbalances is to downsample, which is to draw a subset of the dataset such that the samples are evenly distributed. Another approach is to 

% \begin{itemize}
%     \item  experiment 1: use of one-vs-rest multioutput classification approach, assumption of independence of relations between entity types and relation types. \todo{make correlation heatmap}
%     \item discuss class imbalance issue for relation typing task, solvable without affecting explainability? approaches?
%     \item same as above for entity type prediction task? 
%     \item compare with results of similar experiments? 
%     \item what does it mean that MRR is higher than MAP@k=1 in Task 2 relation typing? 
%     \begin{itemize}
%         \item the predictions are wrong, but at the minimum the correct answer is near the top of the set of each prediction?
%         \item if so, what does that mean with regards to the notion of each model capturing KG triple relations? 
%     \end{itemize}
% \end{itemize}

\section{Future work\label{sec:futurework}}

% \todo{I grabbed this from the discussion part, maybe it has its place here?}
%We note that there is a significant class imbalance in our relation typing task (see \autoref{fig:relation_types}, \autopageref{fig:relation_types}). A common approach to address class imbalances Would be to downsample, which is to draw a subset of the dataset such that the samples are evenly distributed. Another approach is to generate new examples.

% \todo{redact this}
% list of things we've talked about (steps to probe the `unexpected' result of the concat model for relation prediction task, before a more conclusive assessment of the potential for moving to more complex approaches to merge the information in contextual word embeddings and KG embeddings)

The nature of our experiments being of a proof of concept, we have many projected improvements and projects for further research. The majority of our work has been setting up the framework for the two tasks: preparing the data and establishing the evaluation metrics. The areas of improvement on the tasks we have carried out are outlined below:

\begin{enumerate}
    %\item 
    \item applying PCA on the embeddings in order to test for the \blockquote[\cite{bellman2015adaptive}]{curse of dimensionality};
    \item using a simple, but higher-capacity, model such as MLP;
    \item adapting our model in order to prefer nominals as head words \cite{choi2018ultra};
    \item analyse the interpretability of the relation prediction task and the effect of linguistic features such as sentence structure in our setting.
\end{enumerate}

We have made the hypothesis based on our results of the second experiment that, perhaps dimensionality could be an explaination that the scores were lowest for the concatenated input of ELMo embeddings and knowledge graph embeddings. In order to test this hypothesis and further explain the results of our experiments, we propose further processing the input via principal component analysis in order to reduce the dimensionality, presumably resolving this possible source of error. Although we have privileged interpretability over performance in our experiments, it could be useful to test other, higher-capacity models such as MLP in order to see if our hypothesis holds. 
Also, the metrics during the evaluation are fine-grained, so an analysis taking into consideration a coarser baseline (such as raw precision) may contribute to a holistic analysis of each of the embedding model's performance. 

In addition, we have made some suggestions for further experiments along the line of a linguistic interpretation of our results. Linguistically, we presume that nominals carry information about an entity mention. In fact, in a similar entity-focused experiment setting, \cite{choi2018ultra} has privileged selecting common nouns which are present in multi-word entities in order to capture type information: \textquote{Many  nominal  entity  mentions  include  detailed type  information  within  the  mention  itself. [Nominal entity mentions] provide a somewhat noisy, but very easy to gather, context-sensitive type signal}. We therefor propose prioritizing selection of common nouns during pre-processing for task 1. Finally we propose an analysis of linguistic aspects that may provide explaination for results of our second task.

As previously mentioned in \autoref{subsec:Augment}, there are no currently established methods for jointly embedding ELMo contextualized embeddings with knowledge graph embeddings. While staying within the framework of ELMo and the possible integration of Freebase facts, we propose further methods in terms of feasible ways to study contextualized embeddings. It would be possible to learn a mapping from entities' ELMo embeddings to the corresponding KG embeddings. Mapping embeddings to knowledge embeddings has been previously experimented, as an enquiry into the interpretability of what information ``static'' embeddings hold \cite{camacho2018word}.  Similar experiments have been achieved by mapping embeddings to definitions. Such an enquiry would be very interesting to pursue \cite{chang2019does}.

% use transformer networks (so attention over the embeddings) to create the combined embedding, to analyse which aspect of each embedding is used for what

%    \item error analysis for Task 2 relation typing: what are the predictions (i.e. higher-rank predictions) before the correct prediction is made? is the gold label  similar/dissimilar to these wrong higher-ranked relations? If dissimilar, it suggests that the emb model is not suited for relation prediction, if similar, it could be possible the emb model is not capturing sufficient information to `disambiguate' between these relations

\section{Conclusion\label{sec:conclusion}}
%\todo{postponed}

In conclusion, we provide two downstream tasks with datasets applicable equally on contextual embeddings, KB embeddings and models using both textual information and information from KBs. 
Those tasks are typical and generic downstream tasks enabling the estimation of the performance of embedding models on a wide range of applications.

Also, by using contextual and KB embeddings and combining them in a new embedding, we explore a new approach at modelling KB and contextual information on the same level.
In particular, with such a simple embedding model we manage to obtain conclusive results for our typing task, demonstrating the interest of using KB and contextual information as input for embedding.
While we do not manage to obtain conclusive results for our relation prediction task, we have some hypotheses as to why we have such results and ways to explore them.

There is still a lot to do in this field of trying to bridge the gap between knowledge and text processing, and we hope our work can serve as a basis for models and datasets following the same train of thought. % lot to do in this emerging set of methods

%In our work, we explore the possibility of obtaining multi word-sense representations and sense induction to embedding spaces by combining contextual word embeddings and KB embeddings. 
%downstream applications???
%Previous work on obtaining sense representations falls under two distinct clusters: unsupervised methods and supervised resource-specific methods.

% new dataset adapted to tasks
% concat (side hypothesis)

%   we have conclusive results for task typing
%   we do not have conclusive results for task relation & we have hypothesis as to why it is not conclusive
% lot to do in this emerging set of methods

% what we've done: create datasets, select two tasks, 

\newpage
\bibliographystyle{abbrv}  
\bibliography{references}

\end{document}